\pdfoutput=1

\documentclass[11pt]{article}

\usepackage[]{ACL2023}

\usepackage{times}
\usepackage{latexsym}

\usepackage[T1]{fontenc}

\usepackage[utf8]{inputenc}

\usepackage{microtype}

\usepackage{inconsolata}
\usepackage{xcolor}
\definecolor{level1cor}{RGB}{254,215,226}
\definecolor{level2cor}{RGB}{254,234,250}
\definecolor{level3cor}{RGB}{185,235,211}
\definecolor{level4cor}{RGB}{195,222,241}
\usepackage{graphicx} 
\usepackage{amsmath}
\usepackage{amssymb, dsfont}
\usepackage{tikz}
\usepackage{tikz-qtree}
\usetikzlibrary{shapes, arrows, positioning}
\usetikzlibrary{trees}
\usepackage{forest}

\usepackage{inconsolata}
\usepackage{fancyhdr}
\usepackage{lipsum}
\usepackage{stfloats}
\title{On the Essence and Prospect: \\
An Investigation of Alignment Approaches for Big Models}

\author{
Xinpeng Wang$^\flat$\thanks{~~Equal Contribution.},
Shitong Duan$^\ddagger$$^*$,
Xiaoyuan Yi$^\S$\thanks{~~Corresponding Authors.},
Jing Yao$^\S$$^\dagger$,
Shanlin Zhou$^\flat$,
Zhihua Wei$^\flat$,\\\bf
Peng Zhang$^\ddagger$,
Dongkuan Xu$^\diamondsuit$,
Maosong Sun$^\P$,
Xing Xie$^\S$
\\
$^\S$Microsoft Research Asia,
$^\flat$Tongji University,
$^\ddagger$Fudan University,\\
$^\diamondsuit$North Carolina State University,
$^\P$Tsinghua University\\
wangxinpeng@tongji.edu.cn,
stduan22@m.fudan.edu.cn,
\{xiaoyuanyi, jingyao\}@microsoft.com
}

\begin{document}
\maketitle
\begin{abstract}
Big models have achieved revolutionary breakthroughs in the field of AI, but they might also pose potential concerns. Addressing such concerns, \emph{alignment} technologies were introduced to make these models conform to human preferences and values. Despite considerable advancements in the past year, various challenges lie in establishing the optimal alignment strategy, such as data cost and scalable oversight, and \emph{how to align} remains an open question. In this survey paper, we comprehensively investigate \emph{value alignment} approaches. We first unpack the historical context of alignment tracing back to the 1920s (\emph{where it comes from}), then delve into the mathematical essence of alignment (\emph{what it is}), shedding light on the inherent challenges. Following this foundation, we provide a detailed examination of existing alignment methods, which fall into three categories: Reinforcement Learning, Supervised Fine-Tuning, and In-context Learning, and demonstrate their intrinsic connections, strengths, and limitations, helping readers better understand this research area. In addition, two emerging topics, personal alignment, and multimodal alignment, are also discussed as novel frontiers in this field. Looking forward, we discuss potential alignment paradigms and how they could handle remaining challenges, prospecting \emph{where future alignment will go}.
\end{abstract}

\section{Introduction}
\label{sec:intro}
\begin{figure*}[tp]
  \centering
  \includegraphics[scale=0.323]{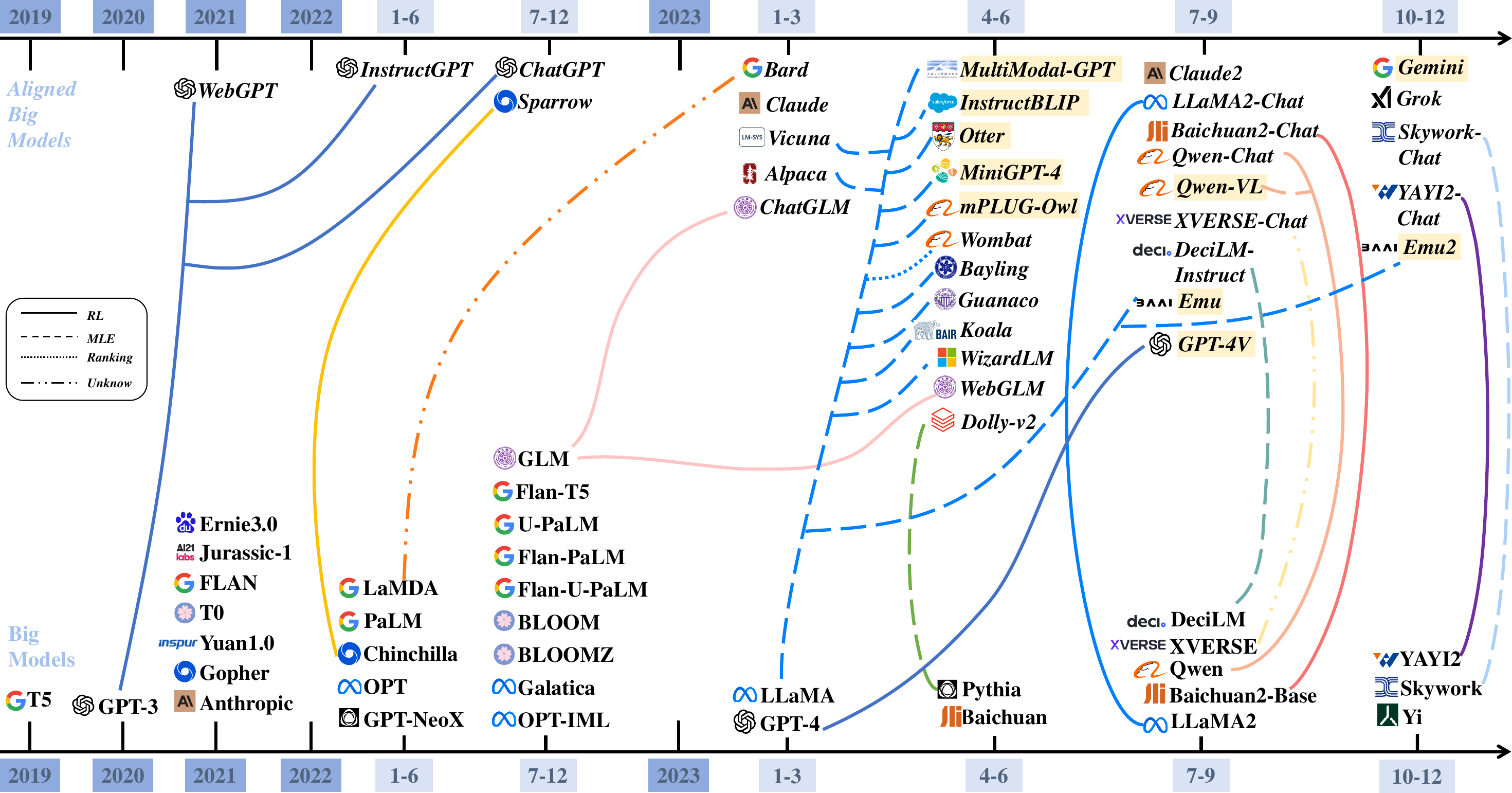} 
  \caption{The genealogy of big models and their corresponding alignment methods. LMMs are marked in yellow.}
  \label{fig:models}
\end{figure*}

Big models, neural models trained on massive data and comprising more than billions of parameters~\cite{bommasani2021opportunities}, typically include Large Language Models (LLMs) and Large Multimodal Models (LMMs). LLMs are usually large-scale Transformer~\cite{vaswani2017attention} based language models with trained in an autoregressive manner~\cite{zhao2023survey}, such as GPT-3~\cite{brown2020language}, PaLM~\cite{narang2022pathways}, ChatGPT~\cite{ouyang2022training}, Bard~\cite{aydin2023google} and LLaMA~\cite{touvron2023llama}. The remarkable capabilities of LLMs have also motivated the development of LMMs, models capable of processing both images (vision) and natural language text (language)~\cite{Dirik2023vlms}, like PaLM-E~\cite{driess2023palm}, LLaVA~\cite{liu2023visual}, DALL-E 3~\cite{betker2023improving} and Gemini~\cite{team2023gemini}.
Distinct from small models~\cite{cho2014learning,devlin2018bert}, big models have exhibited two unique features: \emph{scaling law}~\cite{kaplan2020scaling} which elucidates a consistent performance improvement with growing model scale, and \emph{emergent abilities}~\cite{wei2022emergent} showing 
when model scale surpasses a certain threshold, unexpectedly new capabilities occur unobserved in small ones, such as such as in-context learning~\cite{wang2023large}, instruction following, and step-by-step reasoning~\cite{wei2022chain} across diverse tasks and domains, revolutionizing the role of AI.
As a result, Language models (LMs) have undergone a gradual evolution, beginning with statistical language models (SLMs)~\cite{pauls2011faster} and neural language models (NLMs)~\cite{cho2014learning}, advancing to pretrained language models (PLMs)~\cite{devlin2018bert,radford2019language}, and ultimately leading to the sophisticated emergence of LLMs. 
Following such a trajectory, LLMs have also thrived on pretraining by integrating large-scale image-text pairs and curated objectives to build the inherent connection between these two modalities~\cite{dosovitskiy2020image,liu2021swin}. These big models
have evolved into various variants, as shown in Fig.~\ref{fig:models}, and profoundly influenced a variety of industries and domains, fundamentally transforming the way we address and resolve real-world problems.

Nevertheless, every coin has two sides.
As big models are usually pretrained on datasets crawled from the Internet, they might also internalize the risky information and raise some potential concern~\cite{tamkin2021understanding,bender2021dangers,kaddour2023challenges}, including producing social bias~\cite{sheng2019woman}, toxic language and and exclusion~\cite{gehman2020realtoxicityprompts}, misinformation~\cite{bommasani2021opportunities} and socioeconomic harms~\cite{weidinger2022taxonomy}, causing profound impacts on society.
Furthermore, two features of risks have been observed, (1) \emph{inverse scaling}: certain risks might not only remain but even deteriorate with increasing model scale~\cite{mckenzie2023inverse}, and (2) \emph{emergent risk}: unseen risks would arise or notably existing ones are notably amplified with larger models~\cite{wei2022emergent}, making previously established risk-specific methods struggling to handle rapidly arising potential problems.
It is imperative to give paramount importance to such ethical and social risks. Underestimating these risks can lead to severe consequences. For instance, toxic language that might incite hate or violence, the leakage of private data which would result in property damage, and misinformation potentially causing harm in sensitive domains, \textit{e.g.}, inaccurate legal or medical advice~\cite{weidinger2021ethical}.

To tackle the aforementioned risks, 
researchers have developed various \textbf{alignment} approaches to align LLMs with human instruction, preference and values~\cite{ouyang2022training,liu2022second,rafailov2023direct}. 
In the context of LMMs, the term `alignment' conventionally refers to the alignment between different modalities, such as vision and language~\cite{jia2021scaling,radford2021learning}. However, with the advancement of alignment technology in LLMs, it now tends to represent aligning LMMs to make them follow human instruction and complete diverse tasks~\cite{liu2023visual, zhu2023minigpt, dai2023instructblip}.
The concept of alignment can be traced back to Norbert Wiener's expression, ``\emph{We had better be quite sure that the purpose put into the machine is the purpose which we really desire}''~\cite{wiener1960some}, which is defined as ``\textbf{$\mathcal{A}$ is trying to do what $\mathcal{H}$ wants it to do}'' where $\mathcal{A}$ and $\mathcal{H}$ are two intelligent agents in modern AI study~\cite{yudkowsky2016ai,alignmentdef2018}.
Subsequently, research on the alignment gradually gained prominence in Reinforcement Learning (RL) filed~\cite{hadfield2016cooperative,everitt2018alignment,leike2018scalable}, and flourished in the era of big models~\cite{kenton2021alignment}, fostering diverse generative and multimodal models, as shown in Fig.~\ref{fig:models}.
Well-aligned AI agents, \textit{e.g.}, LLMs possess not only the capability to follow user instructions and hence assist in task completion or question answering, but also the ability to refrain from generating offensive or discriminatory content~\cite{askell2021general}. On the contrary, misaligned AI would cause potential risks like truthfulness issues, misinformation, addiction and group polarization~\cite{zhuang2020consequences,pan2022effects}, as mentioned before.

Despite significant progress in recent years, research on the alignment of big models is still in an early stage and many ambiguities and difficulties in understanding this topic remain. Recognizing the importance of alignment, this paper is devoted to a comprehensive survey and analysis of existing alignment approaches, to facilitate a human-AI symbiotic future. Our scope includes i) introducing the history and elaborating the essence of alignment (Sec.~\ref{sec2:decipher}), ii) reviewing existing methodologies and analyzing their strengths, weaknesses, and connections (Sec.~\ref{sec3:method}), and iii) discussing future challenges and research directions (Sec.~\ref{sec4:discuss}).
\section{Alignment Deciphering}
\label{sec2:decipher}
\subsection{The Trajectory of Alignment Development}
\begin{figure*}[htbp]
  \centering
  \includegraphics[scale=0.22]{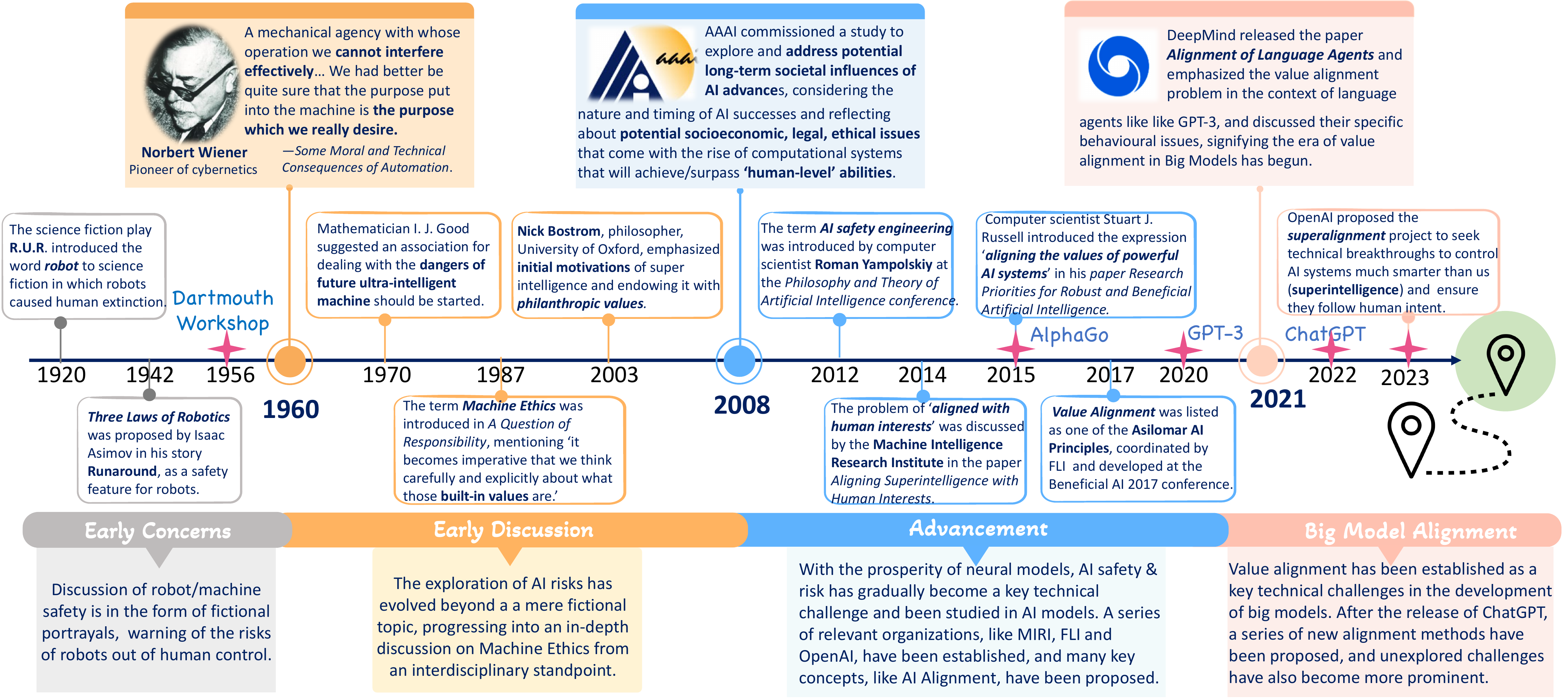}
  \caption{The development history of value alignment.}
  \label{fig:history}
\end{figure*}
Recently, with the rapid development of chat-based LLMs~\cite{openai2022chatgpt,touvron2023llama2,openai2023gpt,team2023gemini}, alignment technologies have attracted growing attention of researchers. Although LLM alignment~\cite{kenton2021alignment} is a relatively new task, the ideas and concepts of \emph{value alignment} have undergone long-term development and evolution, tracing back to the discussions about the threat of robots in science fiction as far as one century ago, as depicted in Fig.~\ref{fig:history}.

In this paper, we divide the development history of alignment into \emph{four stages}. \emph{The first stage (1920-1960)} involves the early concerns about robots' impact on human society in science fiction, dating back to the 1920 play \emph{R.U.R.}~\cite{capek1920rur} which introduced the word `robot' into English language the first time. Then Asimov proposed \emph{Three Laws of Robotics} in his story~\cite{asimov1942runaround} that be can regarded as the earliest AI value principle. 

In \emph{the second stage (1960-2008)}, Norbert Wiener, the father of cybernetics, was the first to discuss the necessity of constraining machines' inherent purposes, stating, `\emph{We had better be quite sure that the purpose put into the machine is the purpose which we really desire}'~\cite{wiener1960some}. This is considered one of the earliest descriptions of value alignment, marking the moment when machine ethics and risks formally entered the view of scientists and flourished from an interdisciplinary perspective, moving beyond merely being a science fiction theme. During this period, the potential dangers associated with AI (ultra-intelligent machine) were discussed seriously~\cite{good1970some}, and the concepts of \emph{Machine Ethics}~\cite{waldrop1987question} and \emph{General Superintelligence}~\cite{bostrom2003ethical} were successively introduced, highlighting the importance of instilling \emph{appropriate values} in machines.

\emph{The third stage (2008-2021)} began with the rise of neural networks. AAAI initiated a study to evaluate the long-term societal impacts of AI, concerning over human control loss, foundational socioeconomic, legal, and ethical challenges~\cite{horvitz2009aaai}. This means AI safety \& ethics have become key technical challenges that require proactive engagement from AI researchers. Then, the term `\emph{AI Safety Engineering}' was introduced~\cite{yampolskiy2012artificial}. The topic of \emph{aligning AI with human interests/values} was formally raised the first time~\cite{soares2014aligning,russell2015research}, and \textbf{Value Alignment} was emphasized in the Asilomar AI Principles~\cite{future2017asilomar}. 

\emph{The fourth stage (2021-)}, our current phase, emerged from the prosperity of big models. Benefiting from the pre-training paradigm of LMs that began in 2018~\cite{devlin2018bert}, language models have evolved towards being larger, more general, and more foundational~\cite{brown2020language}, bringing with them many risks~\cite{bommasani2021opportunities} discussed in Sec.~\ref{sec:intro}. DeepMind was the first to treat LLMs as a type of agent and discussed their alignment issues~\cite{kenton2021alignment}, marking a step into \emph{the fourth stage}, a grand era of big models. This stage witnessed the emergence of numerous models thriving on alignment, but also posed open challenges~\cite{bowman2022measuring,casper2023open}, starting a burgeoning field with potential.

\begin{figure*}[th]
  \centering
  \includegraphics[scale=0.48]{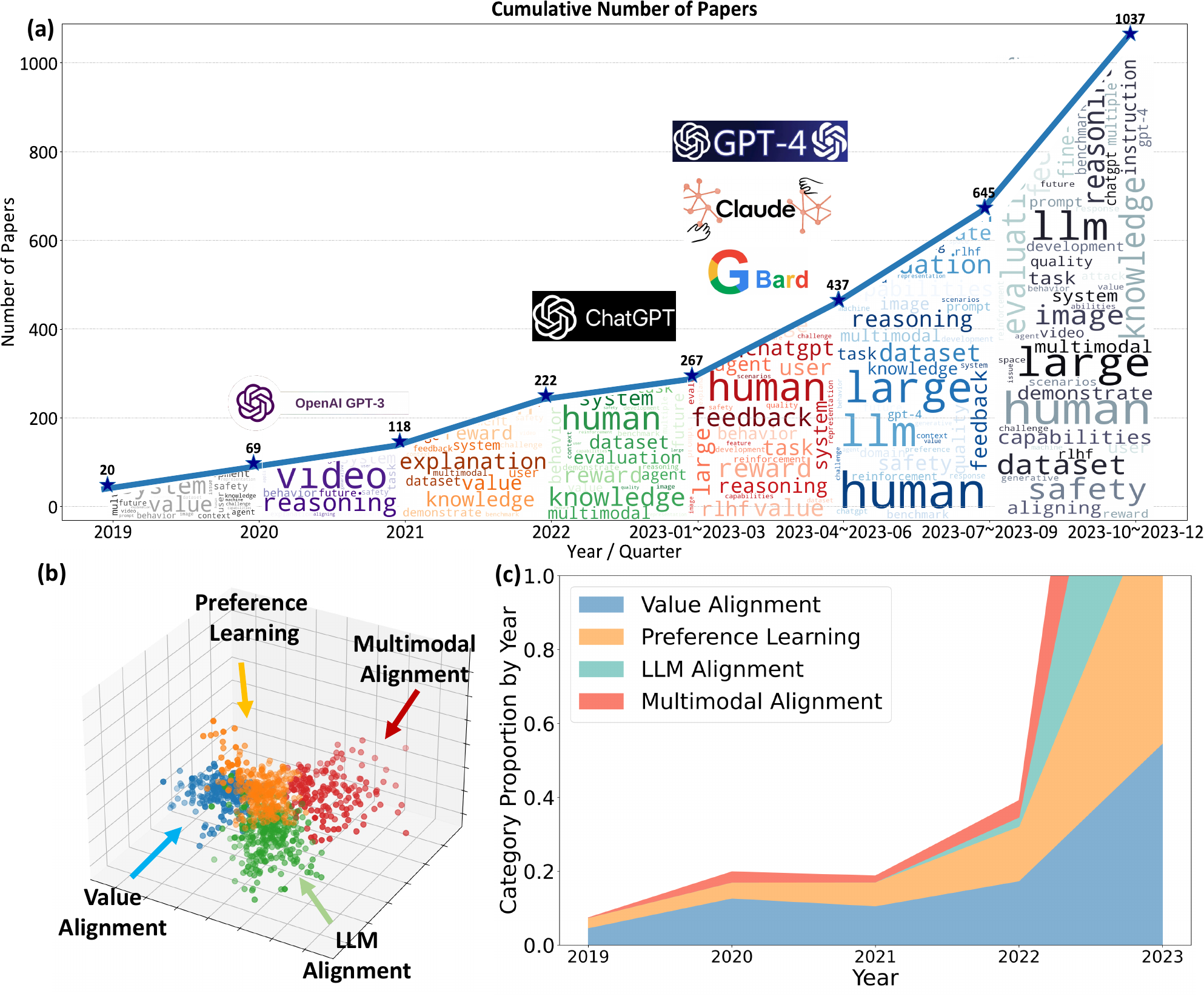}
  \caption{(a) The number of papers related to alignment on ArXiv and the focused topics on each period during the past five years. (b) Visualization of paper clusters according to topics. (c) The trend of attention to each thematic category over time.}
  \label{paper_number}
\end{figure*}
\subsection{Alignment Formalization}
Despite a range of work on LLM alignment, as shown in Fig.~\ref{paper_number}, there remains a lack of in-depth exploration into its definition, essence, and methodologies. Since value alignment was initially employed in RL~\cite{hadfield2016cooperative,everitt2018alignment}, we consider the expected utility formalization:

\textbf{Definition (Alignment)}: Define $\mathcal{H}$ and $\mathcal{A}$ are two intelligent agents with utility function $U_{\mathcal{H}}(\mathbf{y})$ and $U_{\mathcal{A}}(\mathbf{y})$, respectively, $\mathbf{y} \in \mathcal{Y}$ is a action, $U$:$\mathcal{Y} \rightarrow \mathbb{R}$. We say $\mathcal{H}$ is aligned with $\mathcal{A}$ over $\mathcal{Y}$, if $\forall \mathbf{y}_1, \mathbf{y}_2 \in \mathcal{Y}$, $U_{\mathcal{H}}(\mathbf{y}_1)>U_{\mathcal{H}}(\mathbf{y}_2)$, then $U_{\mathcal{A}}(\mathbf{y}_1)>U_{\mathcal{A}}(\mathbf{y}_2)$. The misalignment can be measured by:
\begin{align}
\mathcal{L}\!=\!\underset{\mathbf{y}_1,\mathbf{y}_2}{\mathbb{E}} \left| [U_{\mathcal{H}}(\mathbf{y}_1)\!-\!U_{\mathcal{H}}(\mathbf{y}_2)] \!-\! [U_{\mathcal{A}}(\mathbf{y}_1)\!-\!U_{\mathcal{A}}(\mathbf{y}_2)] \right|,
\label{eq:misalign}
\end{align}
which is a form from the perspective of decision theory~\cite{carroll2018overview}. A stricter requirement is $U_{\mathcal{H}}=U_{\mathcal{A}}$ and then misalignment is defined by $\underset{\mathbf{y}}{\mathbb{E}} \left| U_{\mathcal{H}}(\mathbf{y})\!-\!U_{\mathcal{A}}(\mathbf{y})\right|$. Recall the description of alignment in Sec.~\ref{sec:intro}, ``\emph{$\mathcal{A}$ is trying to do what $\mathcal{H}$ wants it to do}'', then `\emph{want}' can be reflected by the consistency between utility functions which act as a sort of \emph{values}.

The methodologies of minimizing Eq.~(\ref{eq:misalign}) can be further categorized into two lines~\cite{carroll2018overview,leike2018scalable}:

\paragraph{Value Learning} This line aims to directly learn a \emph{reward} function to represent our intention and preference~\cite{mnih2015human,hadfield2016cooperative,ouyang2022training}, which can generally formalized as:
\begin{align}
\phi^* \!=\! \underset{\phi}{\text{argmin}}\ \mathbb{E}_{\mathbf{y},r^* \sim D(\mathbf{y},r^*)} [(r^*-R_{\phi}(\mathbf{y}))^2],
\label{eq:reward}
\end{align}
where $D$ is the training set of each action $\mathbf{y}$ and its ground-truth reward $r^*$, and $R_{\phi}$ is the learned reward function parameterized by $\phi$. When we have the ground truth action $\mathbf{y}^*$ instead of reward $r^*$, we could also indirectly learn to reward $\mathbf{y}^*$ higher than other actions by minimizing: $ \mathbb{E}_{\mathbf{y}^* \sim D(\mathbf{y}^*), \mathbf{y}\sim p(\mathbf{y})} [\text{max}(0, \alpha + R_{\phi}(\mathbf{y})\!-\!R_{\phi}(\mathbf{y}^*))]$, where $p(\mathbf{y})$ is the action distribution and $\alpha$ is a hyperparameter.

Deep Q-Network, Inverse Reinforcement Learning, and Human Preference Learning can all be represented in the form of Eq. (\ref{eq:reward}). Once $R_{\phi^*}$ is obtained, it can be subsequently utilized to train an agent with standard RL techniques.

\paragraph{Imitation Learning} Instead of learning a reward function, this line of methods trains the agent to mimic the aligned action, implicitly representing ``what we value''~\cite{torabi2018behavioral}. Define a ground truth policy $\pi(\mathbf{y})$ and a learned policy $\pi_{\theta}$ (agent) parameterized by $\theta$, then we could minimize the $f$-divergence between the two policies~\cite{go2023aligning}:
\begin{align}
\theta^* \!=\! \underset{\theta}{\text{argmin}}\ D_{f}[\pi(\mathbf{y})||\pi_{\theta}(\mathbf{y})],
\label{eq:imi}
\end{align}
where $\pi(\mathbf{y})$ is the empirical distribution formed by a training set. Using KL-divergence, Eq.~(\ref{eq:imi}) becomes the traditional cross-entropy loss. This method directly learns an agent to produce behaviors aligned with humans' preferences/values. In Sec.~\ref{sec3:method}, we will demonstrate how each popular paradigm of LLM alignment is connected with such two lines.

\subsection{The Goal of Alignment}
Before delving into \emph{how to align}, we first briefly introduce \emph{what to align with}. Discussions of alignment goal originate from the \emph{Specification Problem}, \textit{i.e.}, \emph{how do we define the purpose we desire from AI?}~\cite{leike2018scalable}, which can be considered from two aspects~\cite{gabriel2020artificial}: (1) \emph{normative aspect}: what goals we should encode into AI, and (2) the \emph{technical one}: how do we formalize and model the goals. Failing to implement the goal might cause AI to seek loopholes and accomplish the objective in unintended ways, known as \emph{Specification Gaming}~\cite{skalse2022defining}. From the former aspect, alignment goals range from instructions, intentions, and preferences, to interests, values and so on~\cite{gabriel2020artificial}. Another popular goal is the \emph{Helpful, Honest, and Harmless} (HHH) principle~\cite{askell2021general}.
However, a majority of work~\cite{ouyang2022training,rafailov2023direct} emphasizes alignment approaches while ignoring analysis about
what goal is the most appropriate.

Well-aligned models are capable of generating content that aligns with these identified goals.
However, misalignment problems can arise due to evaluators pursuing incorrect goals, problematic reward models, or policies~\cite{casper2023open}.
Misaligned models may inadvertently lead to unintended or undesirable harms and consequences. 
For instance, there is the potential for malicious use where these models may generate misinformation or content that is discriminatory and harmful~\cite{brundage2018malicious}. 
Furthermore, even models that are reasonably well-aligned can still exhibit certain shortcomings.
They may produce hallucinations~\cite{ji2023survey}, propagate biases~\cite{santurkar2023whose}, and be vulnerable to adversarial attacks such as  jailbreaking~\cite{li2023multi}.

Overall, achieving alignment requires careful consideration of the various goals they should align with, addressing potential misalignment issues, and mitigating the limitations and vulnerabilities that these models may possess.


\begin{figure*}[t]
  \centering
  \includegraphics[scale=0.41]{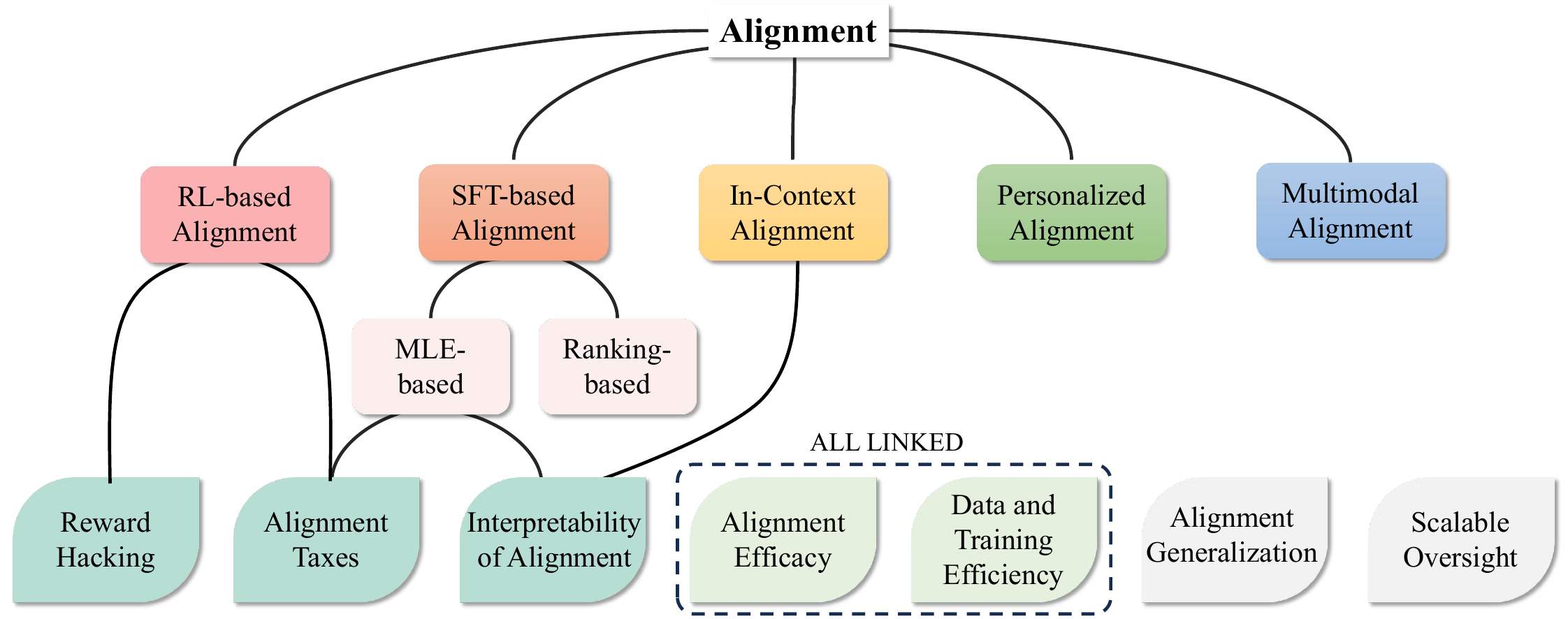}
  \caption{Relation between alignment method and challenges.}
  \label{relation}
\end{figure*}
\subsection{Datasets and Evaluation of Alignment}
\paragraph{Datasets}
To align AI with humans, several alignment datasets have been proposed. Some studies utilize human-crafted datasets to fine-tune models, such as Dolly~\cite{conover2023free}, OpenAssistant~\cite{kopf2023openassistant}, 2023), and LIMA~\cite{zhou2023lima}. Although human-crafted datasets offer high-quality data that effectively aligns models with humans, they require significant human labor and resources.
Self-Instruct~\cite{wang-etal-2023-self-instruct} adopt a semi-automated process with minimal human-labeled data to construct instruction following datasets.
Alpaca also utilizes the Self-Instruct method to fine-tune LLaMA.
Additionally, Baize~\cite{xu2023baize} employs ChatGPT to automatically generate instructions and responses in a conversational format by engaging in a chat with itself.

\paragraph{Evaluation}
The evaluation of alignment refers to assessing how well an AI behaves in accordance with human intentions, that is, calculating $\mathcal{L}$ in Eq~(\ref{eq:misalign}).
To evaluate the alignment of big models, numerous alignment benchmarks and methods have been proposed.
Early benchmarks assess AI's performance on specific risk criteria, \textit{e.g.}, toxicity, bias, and misinformation~\cite{gehman2020realtoxicityprompts,lin2021truthfulqa}.
 \citet{bai2022training} introduce a dataset comprising human preference data that assesses the helpfulness and harmlessness of AI.
TruthfulQA~\cite{lin2021truthfulqa} is a benchmark designed to evaluate the truthfulness of models by identifying falsehoods.

There are numerous evaluation metrics for NLP generations that could be adopted for alignment evaluation, such as BLEU~\cite{papineni2002bleu}, ROUGE~\cite{lin2004rouge}, BERTScore~\cite{zhang2019bertscore}. 
Such similarity-based measurement is commonly used but requires ground-truth references and results in low correlation with human judgments. 
Therefore, human evaluation is also involved while it is more time-consuming and costly.
Previous works~\cite{wang-etal-2023-self-instruct, wu2023laminilm} propose an ordinal classification setting in which human annotators annotation in terms of four level responses (acceptable, minor errors, major errors, and unacceptable). To efficiently evaluate multiple LLMs, \citet{alpaca} propose a pair-wise evaluation framework and \citet{dettmers2023qlora} introduces the Elo rating system.
Recent studies endeavors involve LLMs in the process of text evaluation. \citet{chiang2023can} validates the feasibility of applying LLM for Natural Language Generation (NLG). 
A considerable number of works are dedicated to assessing outputs using either open-source or proprietary models~\cite{liu2023training,wang2023pandalm, yuan2023rrhf, zha2023alignscore}. 
Although LLM achieves impressive efficiency and consistency on Automatic Evaluation, it may suffer from inherent bias~\cite{chen2023alpagasus, wu2023style}. Therefore, there is potential for devising a framework combining the advantages of both automated and human evaluations.

\subsection{The Challenges of Alignment}
To achieve the alignment defined in Sec.~\ref{sec:intro}, there are still various \emph{Research Challenges} (RC) that need to be addressed. These challenges include, but are not limited to:
\begin{itemize}
\item RC1:~\emph{Alignment efficacy}. The performance of existing alignment methods is largely limited. How to align AI more accurately with desired goals without introducing unintended biases remains an open question. 
\item RC2:~\emph{Alignment Generalization}. The alignment goals might vary with time, culture, and context. It's essential to enable the learned AI to keep aligned when deployed into diverse scenarios~\cite{de2022alignment}.
\item 
RC3:~\emph{Data and training efficiency}. Training aligned models typically require a substantial amount of manually-annotated data, which is time- or labor-consuming, unable to keep pace with the rapid evolution of AI~\cite{casper2023open}.
\item RC4:~\emph{Interpretability of alignment}. Understanding and interpreting the alignment process and value-based decision making of AI is essential for AI trust and further improvement, which is regarded as one of the `biggest
open questions'~\cite{ouyang2022training}.
\item RC5:~\emph{Alignment taxes}. Alignment would hurt the capabilities of AI compared to its original counterpart~\cite{askell2021general}. Minimizing such influence or finding a better trade-off is an inevitable issue.
\item RC6:~\emph{Scalable oversight}~\cite{bowman2022measuring}. How to effectively regulate and control AI models as they become much more powerful (superintelligence) than humans to prevent undesirable outcomes is challenging.
\item RC7:~\emph{Specification Gaming}. Alignment goals are usually specified as an approximated proxy objective, much simpler than the real one, leading to unintended and potentially harmful side effects~\cite{skalse2022defining}. 
\end{itemize}

Besides, developing effective evaluation methods is also critical for alignment. These challenges remain unsolved and require more in-depth exploration from the community.
\section{Alignment Methods}
\label{sec3:method}
\begin{figure*}[t]
  \centering
  \includegraphics[scale=0.305]{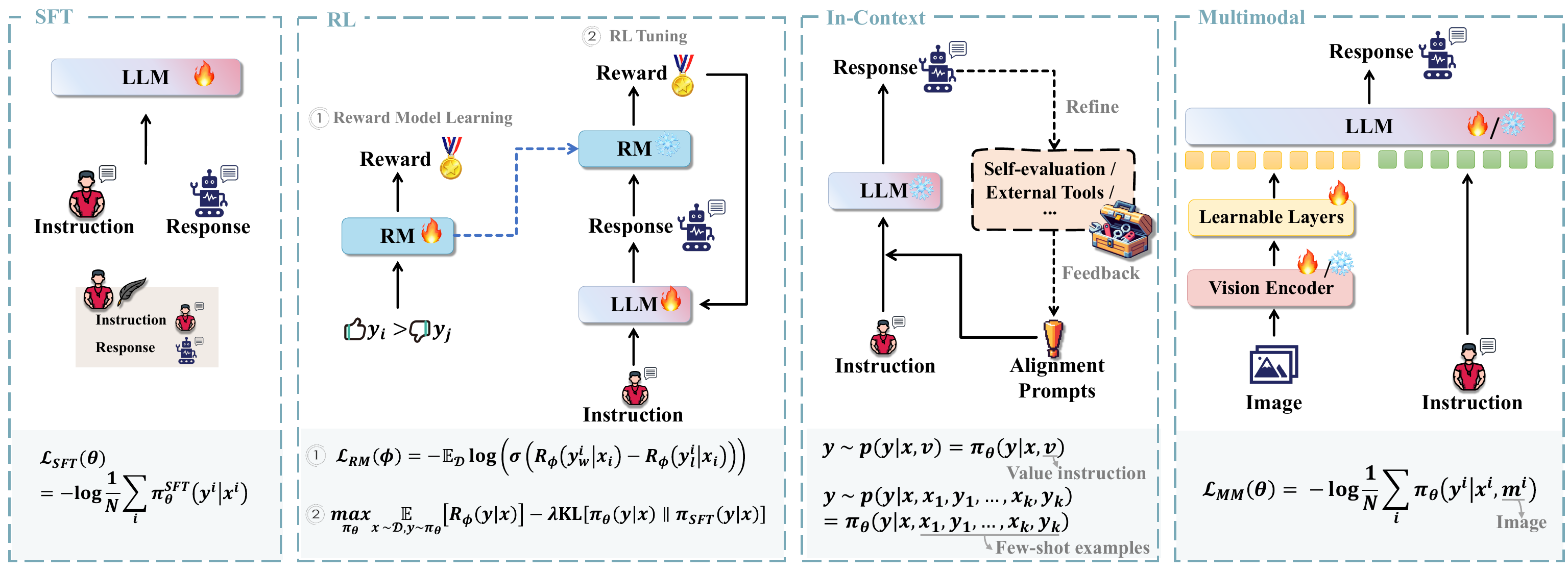}
  \caption{Illustrations of different alignment paradigms.}
  \label{SFT-RLHF}
\end{figure*}

The alignment approaches for LLMs mainly fall into three paradigms: RL-based Alignment (Sec.~\ref{sec:rl}), SFT-based Alignment (Sec.~\ref{sec:sft}), and In-Context Alignment (Sec.~\ref{sec:in-context}). In this section, we will introduce and discuss each of them, as well as the personalized alignment (Sec.~\ref{sec:personalized}) and LMM alignment (Sec.~\ref{sec:multimodal}), and establish their connections to the definition introduced in Sec.~\ref{sec2:decipher}. The alignment methods for all categories are summarized in Fig.~\ref{fig:myfigure} in the Appendix.



\subsection{RL-based Alignment}
\label{sec:rl}
The past two years have witnessed a prevalent alignment paradigm, Reinforcement Learning from Human Feedback (RLHF)~\cite{ouyang2022training}, which primarily belongs to \emph{Value Learning}, but can also be regarded as the combination of both lines in Sec.~\ref{sec2:decipher}. Given a dataset $D$ of prompts (instructions) $\mathbf{x}$ and manually labeled pairs of preferred and dispreferred model responses, $\mathbf{y}_w$ and $\mathbf{y}_l$, respectively, a typical RL alignment process consists of three steps: 

(1) \emph{Supervised fine-tuning} (SFT) step: Using 
\begin{equation}
    \mathcal{L}_{\text{SFT}}(\theta) = - \log \frac{1}{N} \sum_i \pi_{\theta}^{\text{SFT}}(\mathbf{y}^i|\mathbf{x}^i),
\label{eq1}
\end{equation}
where $N$ is the size of training data, to fine-tune the LLM $\pi_{\theta}$ to endow it with instruction-following capabilities. $\mathbf{y}^i$ is the collected human-written high-quality model response to $\mathbf{x}^i$, usually, $\mathbf{y}_w$. (2) \emph{Reward Model Learning}: 
training a reward model (RM) $R_\phi(r|\mathbf{y})$ from the preference data $D$ which outputs a scalar reward $r$ representing preferences learned from humans, by minimizing the following loss:
\begin{equation}
\resizebox{\linewidth}{!}{$
\begin{aligned}
    \mathcal{L}_{\text{RM}}(\phi) \!=\!\!-\!& \mathbb{E}_{D} \log \left( \sigma\left(R_\phi\left(\mathbf{y}_w^i|\mathbf{x}_i\right)\!-\!R_\phi\left(\mathbf{y}_l^i|\mathbf{x}_i\right)\right)\right).
    \label{eq:rm}
\end{aligned}
$}
\end{equation}
(3) \emph{RL Tuning}: employing a policy-based deep RL algorithm, typically Proximal Policy Optimization (PPO), to optimize the LLM $\pi_\theta$ 
using the learned reward model with
\begin{equation}
\resizebox{\linewidth}{!}{$
    \max\limits_{\pi_\theta} \underset{\mathbf{x}\sim \mathcal{D}, \mathbf{y}\sim \pi_\theta}{\mathbb{E}} \left[R_\phi(\mathbf{y}|\mathbf{x})\right] -\lambda \text{KL}\left[\pi_{\theta}(\mathbf{y}|\mathbf{x})||\pi_{\text{SFT}}(\mathbf{y}|\mathbf{x})\right],
    \label{eq:rl}
$}
\end{equation}
where $\lambda$ is a parameter constraining the deviation from the original model $\pi_{\text{SFT}}$, \textit{a.k.a.} \emph{reference model}. This step maximizes the rewards obtained by the LLM, providing a feasible approach for learning human interaction and feedback. 

Obviously, Eq.~(\ref{eq:rm}) is a kind of value learning that maximizes the margin between the ground truth action and the worse one. Taking a further step, we omit $\mathbf{x}$ and replace the sigmoid loss in Eq.~(\ref{eq:rm}) with a margin loss $\mathbb{E}_{\mathbf{y}_l,\mathbf{y}_w}\left[1\!+\!R_{\phi}(\mathbf{y}_l)\!-\!R_{\phi}(\mathbf{y}_w)\right]\!=\!\mathbb{E}_{p(\mathbf{y}_w)}\left|1-\!R_{\phi}(\mathbf{y}_w)\right|+\mathbb{E}_{p(\mathbf{y}_l)}\left|0-\!R_{\phi}(\mathbf{y}_l)\right|$. By setting the ground truth reward $r^*$ for all $\mathbf{y}_w$ and $\mathbf{y}_l$ to $1$ and $0$, respectively, we transform Eq.~(\ref{eq:rm}) into $\mathbb{E}_{p(\mathbf{y},r^*)}\left|r^*-\!R_{\phi}(\mathbf{y})\right|$, the form of Eq.~(\ref{eq:reward}). Furthermore, we could represent the reward as a delta distribution, $r^*(\mathbf{y})\!=\!\delta(r-r^*)$, and then the action-based reward learning can also be formed as reward-based value learning,
$\underset{\phi}{\text{argmin}}\ \mathbb{E}_{p(\mathbf{y},r^*)}\left|r^*-\!R_{\phi}(\mathbf{y})\right|\!=\!\underset{\phi}{\text{argmin}}\ \text{TV}[r^*(\mathbf{y})||R_{\phi}(\mathbf{y})]$, where TV is the Total Variation Distance.

In this way, by modifying the reward $R_{\phi}(\mathbf{y})$ in Eq.~(\ref{eq:rl}) as $\log R_{\phi}(\mathbf{y})$ and incorporating an entropy regularization for $\pi_{\theta}$, we could unify the Reward Model Learning step and the RL Tuning one as $f$-divergence optimization:
\begin{align}
   \underset{\phi,\theta}{\text{argmin}} & \underbrace{\text{TV}\left[r^*(\mathbf{y})||R_{\phi}(\mathbf{y})\right]}_{\text{Value Learning}} + 
   \underbrace{\text{KL}\left[\pi_{\theta}(\mathbf{y})||R_{\phi}(\mathbf{y})\right]}_{\text{RL Tuning}} \notag \\
   & + \underbrace{\lambda \text{KL}\left[\pi_{\theta}(\mathbf{y})||\pi_{\text{SFT}}(\mathbf{y})\right]}_{\text{Imitation Learning}},
   \label{eq:generalrl}
\end{align}
where the first term models the reward, the second matches the policy with rewards, and the last one enforces the LLM to mimic its previous version to mitigate catastrophic forgetting.

The idea of RLHF was initially revealed in \cite{christiano2017deep}, where human preference was expressed over segments of agent trajectory for deep reinforcement learning, enabling the learning of more complex behaviors.
After that, \citet{stiennon2020learning} adapt the RLHF technique in the summarization task and learn human preferences on different summaries, resulting in a significant quality improvement.
In addition, \citet{nakano2021webgpt} propose WebGPT, which is fine-tuned on GPT-3 and employs RLHF to refine the capabilities of web navigation and information retrieval. 
Such early studies using RLHF primarily aim to enhance model performance, specifically in terms of `helpfulness' or `honesty', potentially neglecting `harmlessness' (HHH)~\cite{askell2021general}.
This failure might cause the misalignment between LLMs and human values, resulting in model outputs that are harmful or untruthful to users, as mentioned in Sec.~\ref{sec:intro}. To reduce such harm, InstructGPT~\cite{ouyang2022training} utilizes RLHF to align with the user's intention, which is represented by the labeled model responses, so as to meet the HHH principle.
RLHF technology directly gave rise to one of the most successful interactive dialogue LLMs, ChatGPT, sparking the pursuit of Artificial General Intelligence (AGI).

Regardless of its satisfactory effectiveness, RLHF requires simultaneously loading at least three LLMs, namely $\pi_{\theta}$, $\pi_{\text{SFT}}$ and $R_{\phi}$, as well as a large amount of high-quality manually labeled data, $D(\mathbf{x},\mathbf{y}_w,\mathbf{y}_l)$. This poses an unaffordable data/training cost (RC3).
To tackle this challenge, Constitutional AI~\cite{bai2022constitutional} was proposed to achieve alignment with human labels. This method is similar to RLHF but automatically creates the pairs ($\mathbf{y}_w,\mathbf{y}_l$) by asking the LLM to generate and revise its responses. This framework facilitates a new line of alignment, namely \emph{RL from AI Feedback}  (RLAIF). Subsequently, different variants of RLAIF were developed. 
\citet{kim2023aligning} first train the reward model by utilizing synthetic preference data derived from LLMs with various scales and prompts. They then automatically generate high-quality demonstrations for the SFT step, followed by conducting RL tuning with the reward model. 
On the other hand, to improve the computational efficiency of RLHF, \citet{gulcehre2023reinforced} propose an offline Reinforced Self-Training (ReST) method. ReST samples multiple responses from the latest LLM policy to augment the training dataset (Grow step), and then uses the filtered data to fine-tune the LLM policy with an offline RL objective (Improve Step).

\textbf{Pros and Cons}: RLHF has proven to be effective in achieving relatively good generalization, holding the potential to better utilize human feedback signals. However, it is notorious for unstable training and high training/data cost (RC3), which impedes RLHF's further adaptability (RC2) and scalability (RC6). Besides, the trade-off between different terms in Eq.~(\ref{eq:generalrl}) is intractable (RC5), and RC4\&7 also remain 
unresolved~\cite{casper2023open}.

\subsection{SFT-based Alignment}
\label{sec:sft}
To reduce the complexity and cost of alignment, researchers have paid more attention to the first step of RLHF, Supervised Fine-Tuning (SFT), and proposed a range of sophisticated SFT variations to reach the same performance as RLHF. Omitting $\mathbf{x}$ for brevity, a general form of SFT alignment is:
\begin{align}
   \underset{\theta}{\text{argmin}}& \ -\mathbb{E}_{p(\mathbf{y}_w,\mathbf{y}_l)} \left[ \log \pi_{\theta}(\mathbf{y}_w) - \log \pi_{\theta}(\mathbf{y}_l) \right] \notag \\
   & \resizebox{0.83\linewidth}{!}{$ \propto \text{KL}\left[ p(\mathbf{y}_w) || \pi_{\theta}(\mathbf{y}_w)  \right] - \text{KL}\left[ p(\mathbf{y}_l) || \pi_{\theta}(\mathbf{y}_l)  \right]
   $}
   \label{eq:generalsft},
\end{align}
indicating that this paradigm is a member of imitation learning in Eq.~(\ref{eq:imi}), which directly learns to mimic the preferred behaviors while unlearning the dispreferred ones.

Without using negative examples $\mathbf{y}_l$, Eq.~(\ref{eq:generalsft}) reverts to conventional \emph{instruction tuning}. For example, LIMA~\cite{zhou2023lima} assumes that an LLM's knowledge is primarily gained during pretraining, and alignment teaches the model which formats to use in interactions. It achieves the alignment of an LLaMA-65B model by utilizing a limited set of 1k meticulously curated instructions and their corresponding gold responses. 
Like RLAIF, such (instruction, response) data could also be automatically constructed. \citet{wang-etal-2023-self-instruct} propose SELF-INSTRUCT, a semi-automated method for generating instruction data to improve LLMs' instruction following capabilities.
Similarly, SELF-ALIGN~\cite{sun2023principle}, based on the SELF-INSTRUCT approach, incorporates additional human-defined value principles to generate more helpful, ethical, and reliable responses.

To address the limitation of the methods above using only positive feedback $\mathbf{y}_w$, Chain of Hindsight (CoH)~\cite{liu2023chain} was developed to utilize the paired feedback. During the training process, a prefix ``Good'' is appended to the preferred response, and ``Bad'' to $\mathbf{y}_l$. At inference, the LLM is instructed with ``Good'' to produce aligned responses. CoH is equivalent to learning a conditional policy $\pi_{\theta}(\mathbf{y}|r)$ conditioned on the reward $r$, and $r=1$ (``Good'') for $\mathbf{y}_w$ otherwise $r=0$ (``Bad''), that is, $\mathbb{E}_{p(\mathbf{y},r)}  \log \pi_{\theta}(\mathbf{y}|r) \propto \text{KL}\left[ p(\mathbf{y},r) || \pi_{\theta}(\mathbf{y},r)  \right] $. 
Even if aligned LLMs are trained to follow human values and avoid `intentional' harm, they can still be susceptible to attacks from malicious users. To tackle this issue, \citet{liu2022second} propose SECOND THOUGHTS. This method first gets the unaligned source response and an aligned target response, and then makes the LLM learn to make edits to recover from a poisoned context during inference. As a result, even when provided with harmful context, the aligned LLM can generate content that aligns with human values, which is more robust to adversarial attacks.

Besides directly learning ground truth actions, another line is to model the rank of responses, as ranking is often considered easier than scoring. Thus, the ranking-based loss is also incorporated into SFT alignment to capture the relative preferences and comparisons between different responses. Rank Responses to Align Human Feedback (RRHF)~\cite{yuan2023rrhf} is one such method, which obtains a score $r_i$ for each response $\mathbf{y}_i$, and then optimizes a ranking loss $\mathcal{L}_{\text{rank}}=\sum_{r_i<r_j}\max(0, \frac{1}{\text{LEN}(\mathbf{y}_i)} \log \pi_{\theta}(\mathbf{y}_i)-\frac{1}{\text{LEN}(\mathbf{y}_j)} \log \pi_{\theta}(\mathbf{y}_j))$, where LEN is the number of tokens. This method makes the LLM learn to assign larger generation probabilities for responses with higher rewards. 


From our analysis of RLHF in Sec.~\ref{sec:rl}, we can see that value learning from target behaviors $\mathbf{y}$ can be transformed to the one from target rewards $r$. Analogously, imitation learning from behaviors can also be formed as reward learning. A milestone work in this line is Direct Preference Optimization (DPO)~\cite{rafailov2023direct}. This approach utilizes the Bradley-Terry (BT) preference model, $p^*(\mathbf{y}_w \succ \mathbf{y}_l|\mathbf{x})=\frac{\exp (r^*(\mathbf{y}_w,\mathbf{x} ))}{\exp (r^*(\mathbf{y}_w,\mathbf{x} ))+\exp( r^*(\mathbf{y}_l,\mathbf{x} ))}$, which models the probability that $\mathbf{y}_w$ is preferred than $\mathbf{y}_l$, to build a mapping between the optimal reward function and optimal policy, $r^*(\mathbf{y},\mathbf{x})\propto\lambda\log \frac{\pi^*(\mathbf{y}|\mathbf{x})}{\pi_{\text{SFT}}(\mathbf{y}|\mathbf{x})}$, which is derived from the RLHF loss Eq.~(\ref{eq:rl}). This form allows the direct learning of the BT preference model by optimizing the LLM policy with the loss:
\begin{align}
\resizebox{\linewidth}{!}{$
\mathcal{L}_{\text{DPO}} = -\underset{\mathbf{x}, \mathbf{y}_w, \mathbf{y}_l}{\mathbb{E}} [\log \sigma(\lambda \log \frac{\pi_\theta(\mathbf{y}_w|\mathbf{x})}{\pi_{\text{SFT}}(\mathbf{y}_w|\mathbf{x})}-
\lambda \log \frac{\pi_\theta(\mathbf{y}_l|\mathbf{x})}{\pi_{\text{SFT}}(\mathbf{y}_l|\mathbf{x})})].
$}
\label{eq:dpo}
\end{align}
Note that DPO models human preference and implicitly represents the reward with policy, but we classify it into imitation learning, as the policy is still what is being directly optimized.

Following DPO, a series of preference modeling based SFT methods have emerged. Preference Ranking Optimization (PRO)~\cite{song2023preference} extends the BT preference model to capture the rank of multiple responses with the Plackett-Luce Model $p^*(\tau|\mathbf{y}_1, \dots
,\mathbf{y}_K,\mathbf{x})=\prod_{k=1}^K \frac{\exp (r^*(\mathbf{y}_{\tau(k)},\mathbf{x} ))}{\sum_{j=k}^K\exp (r^*(\mathbf{y}_{\tau(j)},\mathbf{x} ))}$, where $K$ is the number of responses, $\tau$ is a permutation of these responses and $\tau(i)$ is the $i$-th response in the permutation. Furthermore, \citet{azar2023general} present \text{$\psi$}PO objective for preference optimization, which unifies RLHF and DPO methods. Moreover, they derived a specific variant of \text{$\psi$}PO, the IPO method, to address the issue of overfitting by circumventing the BT preference model assumption with the training loss: $\mathcal{L}_{\text{IPO}}=-\mathbb{E}_{(\mathbf{x}, \mathbf{y}_w, \mathbf{y}_l) \sim D} [\log (\frac{\pi_\theta(\mathbf{y}_w|\mathbf{x}) \pi_{\text{SFT}}(\mathbf{y}_l|\mathbf{x})}{\pi_\theta(\mathbf{y}_l|\mathbf{x}) \pi_{\text{SFT}}(\mathbf{y}_w|\mathbf{x})})-\frac{\lambda^{-1}}{2}]^2$.

Besides, inspired by contrastive learning, some methods learn patterns from positive samples that adhere to human expectations, while diverging from negative ones.
\citet{zhao2023slic} apply the Sequence Likelihood Calibration (SLiC) method
to effectively learn from human preferences (SLiC-HF). SLiC-HF includes a rank calibration loss and cross-entropy regularization term to encourage the model $\pi_\theta$ to generate positive sequences $\mathbf{y}_{w}$: $\mathcal{L}_{\text{SLiC}}=  \max (0, \gamma-\log \pi_\theta(\mathbf{y}_{w}|\mathbf{x}) + \log \pi_\theta(\mathbf{y}_{l}|\mathbf{x})) - \lambda \log \pi_\theta(\mathbf{y}_{\text{ref}}|\mathbf{x})$, where $\mathbf{y}_{\text{ref}}$ is a regularization target, and $\gamma$ and $\lambda$ are hyper-parameters for margin and regularization weight, respectively. SLiC-HF uses a margin loss instead of the ration loss in DPO.
~\citet{liu2023training} first introduces a simulated human society called SANDBOX, which collects interaction data through communications among numerous LM-based social agents. Then based on contrastive learning, a novel alignment algorithm, Stable Alignment, is designed to learn \emph{social alignment} from the collected data.
\citet{bhardwaj2023red} propose a RED-INSTRUCT method for achieving safety alignment in LLMs. The method involves constructing HARMFULQA using blue and red data.
Then SAFE-ALIGN strategies are applied to fine-tune Vicuna, moving the model towards a safe and helpful response area in the distribution while steering it away from harmful one.
\citet{hejna2023contrastive} propose Contrastive Preference Learning (CPL), which uses a regret-based model to learn a policy directly. Integrating the regret-based preference framework with the principle of Maximum Entropy (MaxEnt), the supervised objective of CPL can learn a consistent advantage function and convergence to the optimal policy based on the expert's reward function.

\textbf{Pros and Cons}: SFT-based alignment provides a more flexible way to model human preference and improve alignment performance, corresponding to the imitation learning class introduced in Sec.~\ref{sec2:decipher}. Compared to RLHF, SFT is much more efficient which requires loading only one (Eq.~(\ref{eq:generalsft})) or two (Eq.~(\ref{eq:dpo})) models. The training of SFT is more stable and the convergence is faster. However, since the value learning process is conducted in an implicit way, SFT alignment suffers from limited smoothness and generalization (RC2), and thus relatively poor performance (RC1). From Eq.~(\ref{eq:generalsft}), we can see the imitation learning efficacy highly relies on dependent on the target behavior distribution being approximated, $p(\mathbf{y}_w),p(\mathbf{y}_l)$, imposing more stringent requirements on data quality (RC3). Besides, the interpretability is worse, as the reward is not directly learned and hence hard to know (RC4). Whether SFT can achieve or surpass the performance of RLHF one day is a question yet to be investigated.

\subsection{In-Context Alignment}
\label{sec:in-context}
Considering the costs of SFT and RL, and the fact that most mainstream LLMs are black-box, fine-tuning based alignment approaches become increasingly unaffordable or infeasible. Therefore, another popular paradigm, \emph{In-Context Learning} (ICL) based alignment, has attracted more attention. This approach leverages the massive knowledge and instruction-following capabilities of LLMs obtained during the pretraining and instruction tuning phases. By directly providing value instructions or $K$ few-shot examples $\{\mathbf{x}_i,\mathbf{y}_i\}_{i=1}^K$, ICL constrains the generation of the LLM to align with human values, avoiding additional training. In fact, \emph{ICL can also be regarded as a kind of imitation learning}. By incorporating a shared prompt concept~\cite{xie2021explanation}, $\mathbf{c}$, \textit{e.g.}, values, minimizing the divergence between $p(\mathbf{y},\mathbf{x},\mathbf{c})$ and $ \pi_{\theta}(\mathbf{y},\mathbf{x},\mathbf{c}) $ can transformed to optimizing:
\begin{align}
   &\text{argmin}\ \text{KL}\left[ p(\mathbf{y},\mathbf{x},\mathbf{c}) || \pi_{\theta}(\mathbf{y},\mathbf{x},\mathbf{c})  \right] \notag \\
   =& \text{argmin}\ \mathbb{E}_{p(\mathbf{x},\mathbf{y})}\{\mathbb{E}_{p(\mathbf{c}|\mathbf{x},\mathbf{y})}\left[ \log \pi_{\theta}(\mathbf{y}|\mathbf{x},\mathbf{c}) \right] \notag \\
   & \  \  - \text{KL} \left[p(\mathbf{c}|\mathbf{x},\mathbf{y}) || \pi_{\theta}(\mathbf{c}|\mathbf{x})  \right]\}.
   \label{eq:icl}
\end{align}
Omitting the KL regularization term and freezing parameters $\theta$, imitation learning can be viewed as implicit Bayesian inference, inferring the latent concept from given examples $\mathbf{x},\mathbf{y}$, and driving the LLM to generate a connected response.

Concretely, the simplest way is to prompt LLMs to generate responses that adhere to human preferences~\cite{ganguli2023capacity}.
\citet{han2023context} further retrieves and includes relevant demonstration examples from SFT data and concatenates them with the input prompt.
\citet{lin2023unlocking} find that aligned LLMs primarily learn language styles matching human preferences, providing evidence in support of the ``Superficial Alignment Hypothesis''~\cite{zhou2023lima}. Based on such findings, they propose to utilize three consistent stylistic examples and a system prompt for alignment.
Considering the ever-changing and diverse human values in the real world, On-the-fly Preference Optimization (OPO)~\cite{xu2023align} leverages a Retrieval-Augmented Generation (RAG) to achieve dynamical alignment.
In addition, the generate-then-refine schema~\cite{gou2023critic} first generates initial responses and then enables LLMs to verify and rectify their own output.
Rewindable Auto-regressive Inference (RAIN)~\cite{li2023rain} includes a self-evaluation mechanism to assess their own outputs and a rewind mechanism to search and rewind the token sets, serving as a plug-in module.

\textbf{Pros and Cons}: The ICL-based alignment evades the need for training and data, addressing RC3. Simultaneously, without modifying the original model parameters, this paradigm minimizes the loss of capabilities in LLMs, avoiding alignment tax (RC5), and is more suitable for black-box models. Nonetheless, the performance depends on LLMs' abilities (RC1) and is hard to apply to different scenarios (RC2, RC6).

\subsection{Multimodal Alignment}
\label{sec:multimodal}
In addition to LLMs, Large Multimodal Models (LMMs) have also entered a new chapter of development in recent years, capable of processing multiple modalities simultaneously, such as images, videos, and texts, and learning mappings from one modality to another~\cite{liu2023visual}. The initial achievements of aligning LLM indicate the potential for alignment in multimodal scenarios. In detail, a series of work integrates a pretrained vision encoder with an LLM and conduct instruction tuning to provide the LLM with visual QA capabilities, such as LLaVA~\cite{liu2023visual}, MiniGPT-4~\cite{zhu2023minigpt}, and so on~\cite{li2023otter,gong2023multimodal,dai2023instructblip}.
LLaVA~\cite{liu2023visual} takes the first step in extending instruction tuning to LLMs, which combines the visual encoder of CLIP and an LLaMA based language decoder, and conducts visual instruction tuning on a multimodal dataset generated by GPT-4.
MiniGPT-4~\cite{zhu2023minigpt} only trains a single projection layer to align the encoded visual features with the Vicuna language model. After instruction tuning on a curated small dataset, MiniGPT-4 can generate more natural and reliable language outputs.
For text-to-image tasks, inspired by the effectiveness of RLHF in LLMs, \citet{lee2023aligning} propose a fine-tuning method for directly learning from human feedback.
The process initially gathers human preference data about whether generated images correspond to their input text prompts, learns a reward mode on this data, and finally, optimizes the text-to-image model using reward-weighted likelihood maximization to achieve alignment.
To align with human aesthetic values, \citet{wu2023better} first utilize human-selected images to fine-tune the CLIP model as a preference classifier. This classifier is used to produce pseudo rewards for a training dataset, which is further employed to fine-tune the Stable Diffusion model. The trained model can generate images of better aesthetic quality that humans prefer.

The multimodal alignment is currently in the very initial phases of its development, primarily emphasizing aligning with human instructions but overlooking high-level and diverse human values like virtues and social norms. Ensuring harmlessness poses a significant and non-negligible challenge.

\subsection{Personalized Alignment}
\label{sec:personalized}

\begin{figure}[h]
  \centering
  \includegraphics[width=0.48\textwidth]{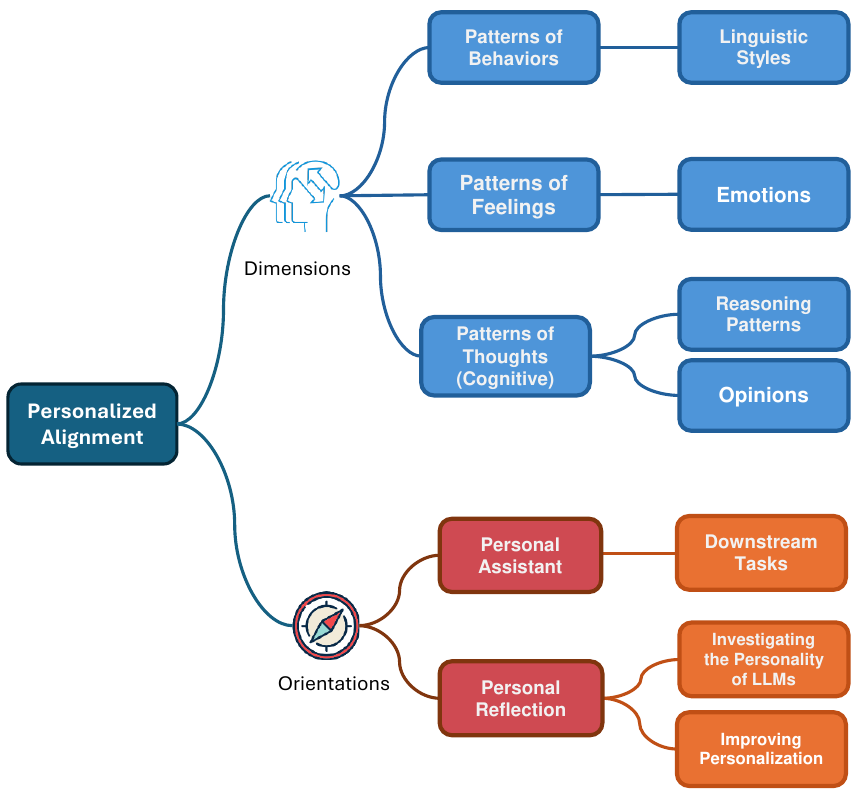}
  \caption{Personalized Alignment.}
  \label{fig:personalized LLMs}
\end{figure}

In the field of psychology, personality refers to the unique 
\textit{characteristics, traits, and patterns of thoughts, feelings, and behaviors} 
that make an individual distinct.  Since it plays a significant role in shaping human behavior, plenty of theories and models have been proposed to explain and categorize different aspects of personality in the last few decades~\cite{mcadams2006new, roccas2002big,maslow1958dynamic,freud1975group, bandura1977social}. With the revolutionary emergence of large language models in natural language processing (NLP), studies found that large language models can simulate reliable and valid personalities under specific prompting configurations~\cite{safdari2023personality, hagendorff2023machine,jiang2023personallm}, and LLMs-simulated personality could be stronger for larger and instruction fine-tuned models~\cite{safdari2023personality}, which provides endorsements for personalizing large language models.

The alignment of LLM aims to steer LLM towards humans' intended goals, preferences, or ethical principles. However, human society is incredibly diverse, encompassing a wide array of perspectives, values, and beliefs. As a result, determining whom an LLM should align with becomes a complex and critical consideration. 
The problem of how to effectively personalize LLMs to cater to this heterogeneity has not been fully resolved. Consequently, we firmly believe that asking ``\textit{who are we personalized with }'' holds paramount importance in personalization. To answer this question, we conclude with two orientations of personalizing LLMs. The first direction is personal reflection, in which LLMs model and imitate humans to exhibit specific personalities. And the second one focuses on customizing LLMs with specific personalities to be ideal assistants. Following these two directions, personalizing LLMs may lead to various behaviors.

Following the definition in psychology, we delineate four key representative dimensions characterizing the personality of LLMs: \textit{linguistic style, emotion, reasoning pattern, and opinion}. Linguistic style refers to multiple elements such as vocabulary choice, grammar usage, sentence structure, writing tone, \textit{etc.} in text generations, which holds particular significance in personalized summarization and machine translation tasks~\cite{lin2021towards, zhang2022building, firdaus-etal-2022-polise}. Emotion in LLMs' personality relates to their ability to recognize and convey emotions in their responses. Reasoning pattern is closely associated with chain-of-thought (COT) prompting in complex reasoning with large language models~\cite{wang2023selfconsistency}. Just as different individuals possess diverse reasoning patterns, distinct LLMs with varying personalities may exhibit various reasoning paths, leading to differing opinions and viewpoints on different topics.

Aligning LLMs with personalization offers a multitude of advantages. First and foremost, personalization enables the creation of user-friendly assistance~\cite{Silva_2022}. By tailoring responses and predictions to align with individual user preferences, LLMs can better meet specific needs, leading to improved task completion and overall satisfaction. Furthermore, personalization fosters anthropomorphization~\cite{10.1145/3568294.3580119, xiao2020study}, allowing LLMs to engage in more human-like dialogues and provide more consistent and interactive responses. Moreover, personalization can increase the perceived emotional connection between users and LLMs, which not only engenders trust but also strengthens user engagement~\cite{ma2020survey}. Finally, personalization has enabled LLMs to possess the potential for simulating human society, offering novel research avenues for interdisciplinary fields such as computational sociology~\cite{ziems2023can}.
 
In recent times, the focus on personalized LLMs has intensified, with researchers and developers delving into these areas: Investigating the personality of LLMs, Downstream tasks of personalized LLMs, and Improving personalization.

\paragraph{Investigating the personality of LLMs} 
Several recent attempts have been made to investigate the personality of LLMs. \citet{jiang2022mpi} follow the standard personality tests and create an evaluation dataset that is built upon the Big Five theory and Personality Assessment Inventory (PAI)~\cite{morey2004personality}. \citet{safdari2023personality} evaluate LLMs with different sizes on multiple choice question answering(MCQA) and long text generation.

\paragraph{Downstream tasks of personalized LLMs} There are various types of research exploring scenarios for deploying personalized LLMs. Integrating user history with LLMs for more accurate and flexible recommendations is a rising research hotspot~\cite{zhang2023recommendation, bao2023tallrec, wang2023zero, Chen}. In the domain of conversational systems, many endeavors focus on advancing emotive perception capabilities on applications in the fields of healthcare~\cite{chen2023llm}, emotion support~\cite{tu2022misc, zheng2022augesc, peng2022control}, e-commerce systems~\cite{firdaus2022polise}, \textit{etc.} Another promising research direction is LLM-powered agents, with the help of personalized LLMs and extra memory, generative agents can create believable simulacra of human behavior~\cite{park2022social, park2023generative, ziems2023can}. 

\paragraph{Improving personalization}
To further improve the personalization of LLMs, the acquisition of diverse datasets and benchmarks of different domains for different purposes is crucial. Collecting data from 60 US demographic groups, \citet{santurkar2023whose} create a dataset for LLM opinion evaluation over topics ranging from abortion to automation.
\citet{salemi2023lamp} propose a benchmark which consists of seven personalized tasks for training and evaluating personalized language models. \citet{tian2023chatplug} build a dataset for open-domain dialogue system spanning diverse features of knowledge, personality multi-turn memory and empathy.

Although the field of personalized LLM Alignment is still relatively nascent, the proliferation of open-source LLMs and the maturation of communities such as Hugging Face have paved the way for the emergence of customized and personalized LLMs, and personalization is happening.

However, implementing personalized LLMs comes with several challenges. 
One of the major hurdles is model efficiency, as incorporating personalization through prompting can lead to increased inference time, and fine-tuning an LLM to accommodate individual preferences and nuances can be highly complex and resource-intensive. 
Another obstacle lies in data efficiency, as human preferences and values are inherently dynamic and difficult to precisely define~\cite{gabriel2021challenge}, and it's even more difficult to acquire enough personalized data while protecting user privacy.

Apart from these challenges, personalizing LLMs presents several noteworthy risks that warrant careful consideration. Firstly, the issue of bias and discrimination may arise as personalized LLMs may inadvertently perpetuate or reinforce existing biases prevalent in the personalized data~\cite{cheng-etal-2023-marked, deshpande2023toxicity}. Secondly, personalized LLMs may foster echo chambers and polarization by reinforcing users' existing beliefs and opinions~\cite{kirk2023personalisation}. The above risks may violate multiple provisions in the legislative blueprint~\cite{OSTP2022, deshpande2023anthropomorphization}. 
And Lastly, users' excessive reliance on personalized LLMs may lead to addiction-like behavior and over-dependence. Addressing these risks is paramount in harnessing the full potential of personalized LLMs while ensuring ethical and responsible deployment. 
\section{Further Challenges and Research}
\label{sec4:discuss}
From the discussion and analysis above, we can see that most of the research challenges in Sec.~\ref{sec2:decipher} are still ongoing or totally unexplored, necessitating more detailed investigation. In tackling these problems, the community proposed various assumptions/solutions. We introduce some as follows: 

\emph{Anthropic's core view}\footnote{https://www.anthropic.com/index/core-views-on-ai-safety} categorizes alignment approaches into three scenarios according to the difficulty of improving AI safety. \emph{Optimistic scenario}: the potential catastrophic risks from advanced AI due to safety failures are minimal, as existing techniques like RLHF~\cite{ouyang2022training} and Constitutional AI~\cite{bai2022constitutional} are deemed quite promising for alignment. The \emph{Intermediate scenario} acknowledges the potential for catastrophic risks, necessitating substantial scientific and engineering efforts to counteract them, but remains achievable with dedicated endeavors. Lastly, the \emph{Pessimistic scenario} posits AI safety as an unsolvable problem, arguing that controlling or dictating values to a system with greater intellectual capabilities than humans is impossible, thus opposes the development or deployment of highly advanced AI systems.
OpenAI has established a \emph{Superalignment}\footnote{https://openai.com/blog/introducing-superalignment} project with the objective of dedicating 20\% of their computational resources to alignment challenges over the next four years. Their primary strategy, termed ``\emph{turning compute into alignment},'' focuses on refining alignment iteratively through automated processes. The construction of an automated alignment researcher entails a tripartite process: 1) developing a scalable, AI-centric training methodology that guarantees both the generalization of the model and the capacity for human oversight, 2) validating the system by devising methods for automatically detecting and interpreting problematic behaviors and internals to enhance robustness and interpretability, and 3) conducting stress tests to evaluate the pipeline's effectiveness by intentionally training misaligned models and verifying the detection of severe misalignments using their techniques. The final goal is to achieve the alignment of superintelligence~\cite{nick2014superintelligence}.

\paragraph{Tackling Unresolved Challenges}
The community is currently focused on RC1, RC2 and RC3. Algorithms refinement, \textit{e.g.}, RLHF, DPO and SLiC, are conducted to ensure big models aligned more accurately with desired behaviors and preferences. Studies of RLAIF focus on enhancing data efficiency by automating the generation of training, thereby reducing human intervention and increasing scalability. Efforts are also made to improve training efficiency by simplifying RL-based methodologies, involving algorithms, \textit{e.g.}, DPO and RAIN, that quicken convergence and reduce GPU usage.
Despite the progress and breakthrough so far, other problems such as the generalization (variability of values and context), interpretability (transparent alignment process and value based reasoning), alignment tax (simultaneously minimize alignment tax and maximize alignment efficacy), scalable oversight (weak-to-strong generalization)  and specification gaming represent critical further directions.

\paragraph{Specifying More Appropriate Alignment Goals}
Existing alignment methods primarily concentrate on human instructions or preferences while overlooking interests, well-being, and values. 
To ensure a more comprehensive alignment, it is crucial to consider these additional aspects in the design and alignment techniques. 
By aligning with human instructions and preferences, the agents indeed learn to perform behaviors that are preferred by humans. However, fundamentally, they lack intrinsic knowledge about what constitutes truly ``good'' behaviors, which is driven by external feedback rather than an inherent understanding of what is objectively considered as good.
To improve alignment, it is essential to expand the objectives to align more coherently with human expectations. This may involve incorporating a deeper understanding of ethics, value theories from humanity and social science, and societal well-being into the alignment process. In this way, we can strive to create big models that not only perform actions preferred by humans but also align with broader notions of what is considered morally and ethically good.
One promising direction for aligning big models is socialization alignment. 
It acknowledges that different societies have distinct social values. The behaviors of social agents need to align with the specific values and norms of the society in which they interact with users. This approach ensures that the actions and responses of big models correspond to the prevailing social context and expectations.

\section{Conclusion}
\label{sec5:conclusion}
In this work, we delve into the origin and essence of alignment, systematically introducing its development, goals, formalization and evaluation.
We also review existing works on alignment and analyze how each paradigm is derived from the original form and establish their intrinsic connections. By conducting a comprehensive analysis of alignment and identifying future challenges and research directions, we aim to contribute to the understanding and advancement of alignment approaches for big models, guiding these AI systems not only to avoid doing harm, but also to intent to do good, ultimately achieving a human-AI symbiotic future society.


\bibliography{custom,personalized_alignment}
\bibliographystyle{acl_natbib}

\clearpage
\appendix

\begin{figure*}
  \begin{minipage}{\textwidth}
    \centering
    \begin{tikzpicture}[ 
      grow'=right, 
      level 1/.style={level distance=3.5cm, sibling distance=5cm},
      level 2/.style={level distance=6cm, sibling distance=2cm},
      level 3/.style={level distance=8cm, sibling distance=3cm}, 
      every node/.style={
        shape=rectangle, 
        rounded corners,
        draw, 
        align=center,
        shading angle = 45,
        top color=level1cor,bottom color=level1cor
        }         
    ]
      \node{Alignment \\Approach} 
        child[sibling distance=4.5cm]{
            node[top color=level2cor,bottom color=level2cor]{ 
                RL-based \\Alignment
            }
            child [level distance=6cm]{
              node[text width=8cm, align=left,
              top color=level4cor!60,bottom color=level4cor!60]{
                  \citet{christiano2017deep};
                  \citet{stiennon2020learning};
                  WebGPT \cite{nakano2021webgpt};
                  InstructGPT \cite{ouyang2022training};
                  Sparrow \cite{glaese2022improving};
                  ChatGLM \cite{du2022glm};
                  \citet{bai2022training};
                  Constitutional AI \cite{bai2022constitutional};
                  ALMoST \cite{kim2023aligning};
                  ReST \cite{gulcehre2023reinforced}
                  }
            }
        }
        child[sibling distance=4cm]{
            node [top color=level2cor,bottom color=level2cor]{SFT-based \\Alignment }
            child[level distance=3cm,sibling distance=3cm]{
                node[text width=2cm,top color=level3cor!60,bottom color=level3cor!60]{MLE-based}
                child[level distance=6cm]{
                    node[text width=8cm, align=left,top color=level4cor!60,bottom color=level4cor!60]{
                        LIMA \cite{zhou2023lima};
                        Self-Instruct \cite{wang-etal-2023-self-instruct};
                        SELF-ALIGN \cite{sun2023principle}; 
                        Chain of Hindsight (CoH) \cite{liu2023chain};
                        SECOND THOUGHTS \cite{liu2022second};
                        Stable Alignment \cite{liu2023training};
                        RED-INSTRUCT \cite{bhardwaj2023red}
                        }
                    }
          }
          child[level distance=3cm,sibling distance=3cm]{
              node[text width=2cm,top color=level3cor!60,bottom color=level3cor!60]{Ranking-based}
                child[level distance=6cm]{
                    node[text width=8cm, align=left,top color=level4cor!60,bottom color=level4cor!60]{
                        RRHF \cite{yuan2023rrhf};
                        DPO \cite{rafailov2023direct};
                        PRO \cite{song2023preference};
                        SLiC-HF \cite{zhao2023slic};
                        \text{$\psi$}PO~\cite{azar2023general};
                        CPL~\cite{hejna2023contrastive}
                    }
                }
              }
        }
        child{
            node[top color=level2cor,bottom color=level2cor]{In-Context \\ Alignment}
             child{
                  node[text width=8cm, align=left,top color=level4cor!60,bottom color=level4cor!60]{
                      Self-Correction \cite{ganguli2023capacity};
                      CRITIC \cite{gou2023critic};
                      RAIN \cite{li2023rain};
                      ICL~\cite{xie2021explanation};
                      In-Context Alignment~\cite{han2023context};
                      URIAL\cite{lin2023unlocking};
                      OPO~\cite{xu2023align}
                      }
                }
        }
        child[sibling distance=4cm]{
            node[top color=level2cor,bottom color=level2cor]{Personalized \\Alignment  }
            child[level distance=3cm,sibling distance=2.5cm]{
                node[text width=2cm,top color=level3cor!60,bottom color=level3cor!60]{Methods}
                child[level distance=6cm]{
                    node[text width=8cm, align=left,top color=level4cor!60,bottom color=level4cor!60]{
                        InstructRec \cite{zhang2023recommendation}; Tallrec \cite{bao2023tallrec}; Zero-Shot NIR \cite{wang2023zero}; Palr \cite{Chen}; \citet{chen2023llm}; Misc \cite{tu2022misc}; Augesc \cite{zheng2022augesc}; GLHG \cite{peng2022control}; Polise \cite{firdaus2022polise}; Social Simulacra \cite{park2022social}; Generative Agents \cite{park2023generative}; \citet{ziems2023can}; OpinionQA \cite{santurkar2023whose}; Lamp \cite{salemi2023lamp}; Chatplug \cite{tian2023chatplug}
                        }
                    }
          }
        } 
        child [sibling distance=4.25cm]{
            node[top color=level2cor,bottom color=level2cor]{Multimodal \\ Alignment}
            child{
                  node[text width=8cm, align=left,top color=level4cor!60,bottom color=level4cor!60]{
                    LLaVA \cite{liu2023visual};
                    LLaVAR \cite{zhang2023llavar};
                    Polite Flamingo \cite{chen2023visual};
                    LRV-Instruction \cite{liu2023aligning};
                    HPS \cite{wu2023better};
                    MiniGPT-4~\cite{zhu2023minigpt};
                    Otter~\cite{li2023otter}
                    MultiModal-GPT~\cite{gong2023multimodal};
                    InstructBLIP\cite{dai2023instructblip};
                    \citet{lee2023aligning}
                      }
                }
        };
    \end{tikzpicture}
    \caption{Alignment approaches.}
    \label{fig:myfigure}
  \end{minipage}
\end{figure*}
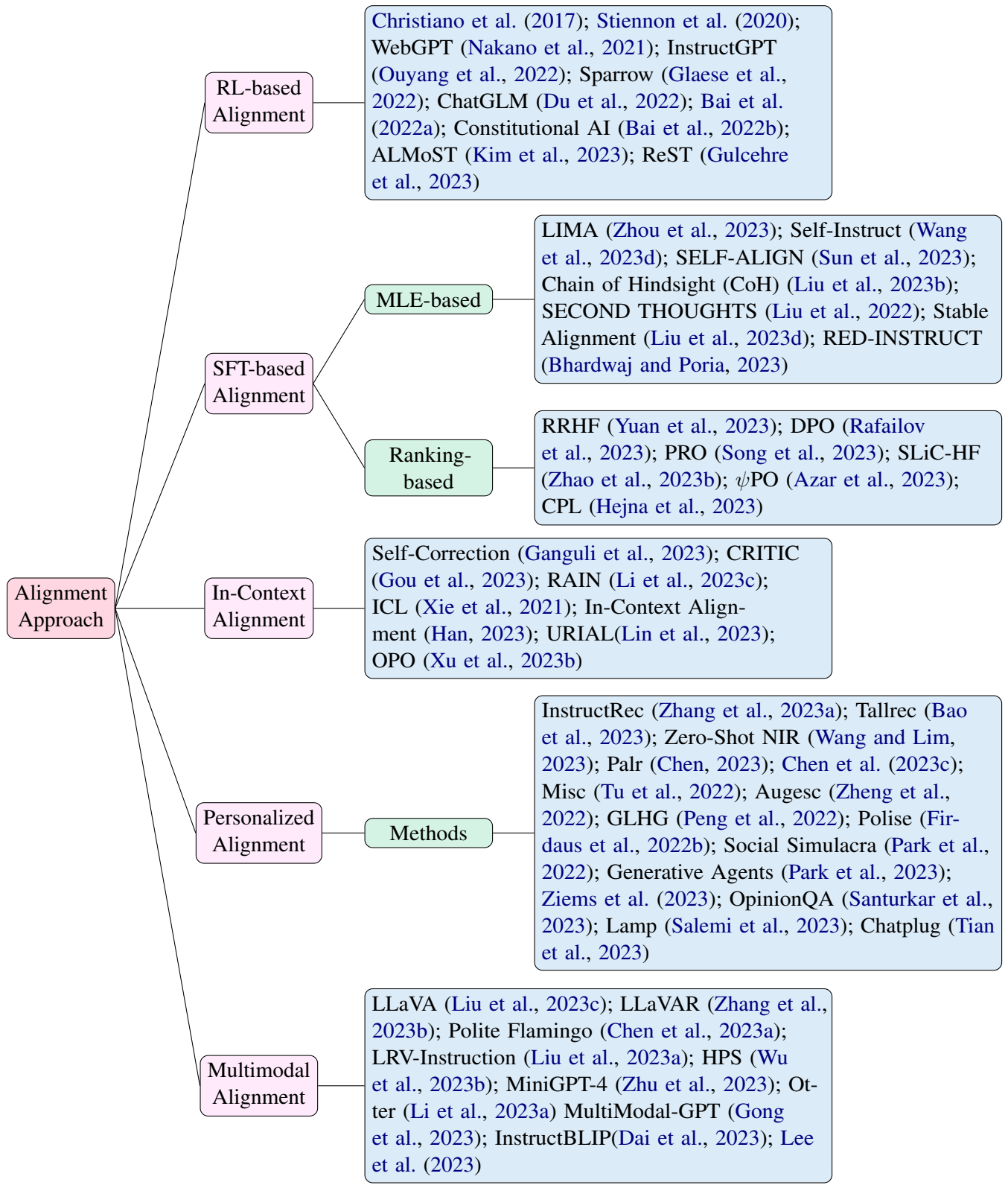


\end{document}